\documentclass[10pt,twocolumn,letterpaper]{article}

\usepackage{wacv}              

\usepackage{colortbl}
\usepackage{xcolor}
\usepackage{multirow}

\colorlet{first}{red!40}
\colorlet{second}{orange!40}
\colorlet{third}{yellow!40}

\usepackage{verbatim} 

\definecolor{wacvblue}{rgb}{0.21,0.49,0.74}
\usepackage[pagebackref,breaklinks,colorlinks,allcolors=wacvblue]{hyperref}

\def\wacvPaperID{1235} 
\def\confName{WACV}
\def\confYear{2027}

\title{EmbedTalk: Talking Head Synthesis using Gaussian Embeddings}

\author{
Arpita Saggar \quad
Jonathan C. Darling \quad
Duygu Sarikaya \quad
David C. Hogg \\ 
University of Leeds, UK\\
{\tt\small \{scasag, j.c.darling, d.sarikaya, d.c.hogg\}@leeds.ac.uk}}

\begin{document}
\twocolumn[{%
\renewcommand\twocolumn[1][]{#1}%
\maketitle
\begin{center}
\vspace{-10pt}
    \centering
    \captionsetup{type=figure}
    \includegraphics[width=0.85\textwidth]{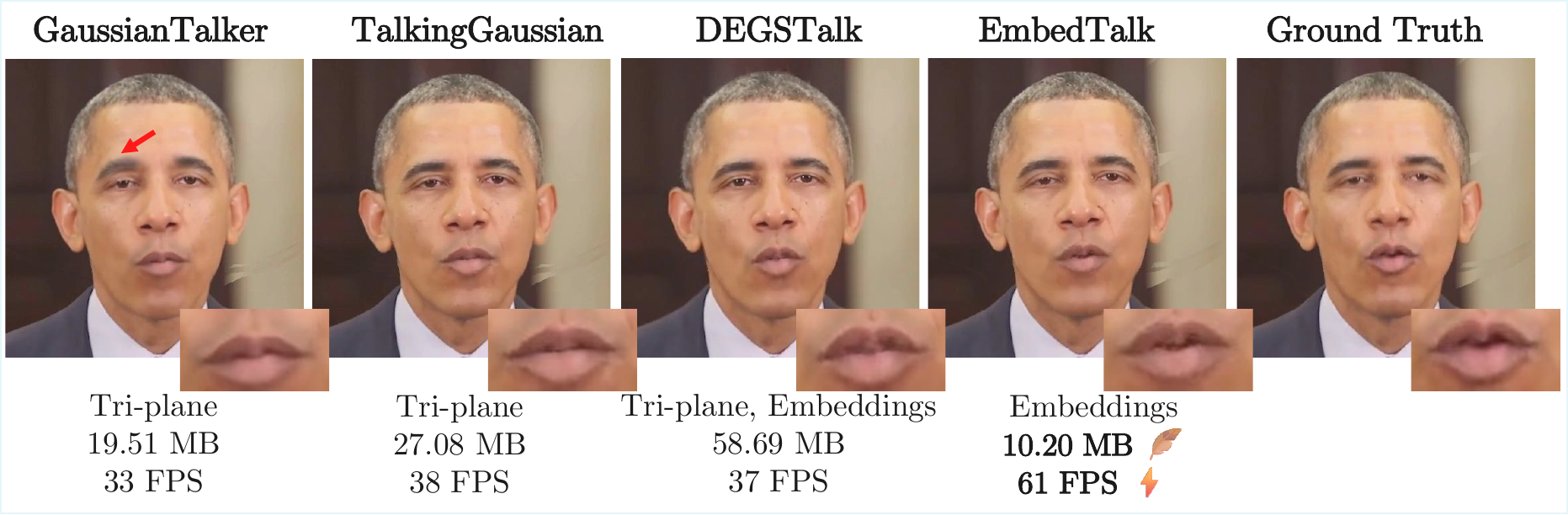}
    \vspace{-9pt}
    \captionof{figure}{EmbedTalk models speech deformations in talking head synthesis with learnable per-Gaussian embeddings, resulting in more accurate mouth
movements and reduced computational overhead compared to tri-plane-based methods \label{fig:embedtalk_teaser}}
\end{center}%
}]
\begin{abstract}
Deformable 3D Gaussian Splatting (3DGS) has emerged as a popular method for real-time talking head synthesis, offering high-quality renderings at low latency. Tri-planes are a common choice for encoding Gaussians prior to deformation since they provide a compact and continuous representation. However, tri-plane encodings are limited by grid resolution and approximation errors introduced by projecting 3D volumetric fields onto 2D subspaces. Recent work has demonstrated the effectiveness of per-Gaussian embeddings for driving temporal deformations in 4D scene reconstruction. We introduce EmbedTalk, which leverages these embeddings to model speech-driven facial deformations for talking head synthesis. Comprehensive experiments show that EmbedTalk improves rendering quality, lip synchronisation, and motion consistency over previous 3DGS-based methods, while remaining competitive with state-of-the-art generative models. Replacing tri-plane features with embeddings also yields significantly more compact models that achieve 60+ FPS on a laptop GPU (RTX 2060 6 GB). Code and further details are available at \href{https://arpita2512.github.io/EmbedTalk/}{https://arpita2512.github.io/EmbedTalk/}.
\end{abstract}

\vspace{-3pt}
\section{Introduction}
\label{sec:intro}

Synthesising real-time talking heads is an important task at the intersection of computer vision and computer graphics, with applicability in film production, teleconferencing, and virtual assistants. The standard pipeline involves deforming a canonical identity representation based on an arbitrary speech sample. Other control signals, such as emotion, gaze direction, or distance to camera may also be used for deformation. The canonical identity is typically represented in either two or three dimensional space. In recent years, methods for synthesising talking heads have been dominated by image-based generative models \cite{alghamdi2022, xu2024vasa, Meng_2025_CVPR, kdtalker, ki2025float} and radiance fields \cite{guo2021adnerf, li2023ernerf, synctalk, degstalk, li2025instag}. Whilst generative models produce high resolution results, they typically do not provide pose control, are slow to render, and can generate uncharacteristic movements which reduce realism and applicability. Within radiance fields, 3D Gaussian Splatting (3DGS) \cite{3dgs} has emerged as the preferred choice due to its fast rendering speed and low memory requirements, compared to neural radiance fields (NeRFs) \cite{mildenhall2021nerf}.

Generating talking heads with 3D Gaussian Splatting involves deforming a canonical Gaussian representation using the speech signal. To encode spatial continuity among the discrete Gaussians, each Gaussian is projected onto 2D subspaces through a tri-plane encoder prior to deformation. \cite{degstalk, li2024talkinggaussian, cho2024gaussiantalker, synctalk_plus, li2025instag, d3talker}. However, tri-plane representations suffer from mirroring artefacts caused by feature entanglement between subspaces \cite{li2025hyplanehead}. Furthermore, using 2D planes to deform a 3D volumetric field introduces approximation errors which subsequently affect audio-visual alignment, as shown in \cref{fig:embedtalk_teaser}. Finally, many previous methods rely on imprecise facial tracking for inferring camera pose, which leads to wobbling around the facial boundary \cite{gaussianheadtalk}.

Prior work has demonstrated the superiority of learnable embeddings for modelling Gaussian deformations in 4D (3D + time) scene reconstruction \cite{ed3dgs}. In this work, we extend the embedding-based deformation paradigm to talking head synthesis. We introduce \textbf{EmbedTalk}, which leverages per-Gaussian embeddings to generate stable talking heads with high audio-visual alignment. To accurately reconstruct high-frequency displacements in the mouth region, we apply positional encodings to the Gaussian embeddings. Spatial coherence is enforced through a local smoothness constraint that encourages similar embeddings for neighbouring Gaussians. Furthermore, we address the head wobbling issue by initialising Gaussians with a stable, dense reconstruction obtained via COLMAP \cite{colmap}. Quantitative and qualitative assessments establish EmbedTalk's superiority in facial fidelity, lip-synchronisation, and motion consistency compared to previous 3DGS-based works. Furthermore, by generating mouth movements that are consistent with an identity's style, our method achieves higher realism than state-of-the-art generative models that often exaggerate motion. Our key contributions are:
\begin{itemize}
    \item A method to leverage learnable Gaussian embeddings for modelling speech-driven facial deformations 
    \item A comprehensive comparative evaluation with recent 3DGS-based and generative methods, covering quantitative assessment, visual comparisons, and a user study 
    \item Ablations and experimental variations validating the efficacy of our design choices 
\end{itemize}

\section{Related Work}

\subsection{Audio-Driven Talking Head Synthesis}

Various approaches have been proposed to generate videos of talking heads conditioned on arbitrary speech samples. Generative models are a popular choice for accomplishing this task since they do not require identity-specific training. Earlier work made use of generative adversarial networks (GANs) to predict deformations in the two-dimensional image space \cite{wav2lip, alghamdi2022}. More recent works replace GANs with diffusion \cite{anitalker, kdtalker, xu2024vasa, echomimicv1, Meng_2025_CVPR} and flow-based models \cite{edtalk, ki2025float}. Some approaches like EDTalk \cite{edtalk} and EchoMimic \cite{echomimicv1} allow explicit pose control through a separate driving video. However, most generative methods are unable to faithfully recreate the identity's speaking style and often produce uncharacteristic motions, leading to reduced realism \cite{li2024talkinggaussian}.

In contrast, approaches based on radiance fields train a separate model for each new identity, enabling generation of personalised talking heads. Many such methods make use of neural radiance fields \cite{mildenhall2021nerf}, with deformations generally driven via a tri-plane representation \cite{Chan2022}. Earlier methods like AD-NeRF \cite{guo2021adnerf} directly conditioned NeRF training on speech input, resulting in slow training and rendering. RAD-NeRF \cite{radnerf} introduced grid-based NeRFs for quicker inference, but its rendering quality was limited by hash collisions. ER-NeRF \cite{li2023ernerf} alleviated this issue through the use of tri-plane encoders, which have since emerged as the de-facto choice for intermediate representations \cite{ye2023geneface, synctalk}. However, most NeRF-based methods are still unsuitable for real-time rendering due to the ray marching step, which requires evaluating the MLP at every sampled point in 3D space. Recently, 3D Gaussian Splatting has emerged as a popular alternative to NeRFs due its fast training and low latency. Similar to NeRF-based approaches, methods that use 3DGS also employ tri-planes to encode Gaussians prior to deformation \cite{cho2024gaussiantalker, degstalk, synctalk_plus, li2024talkinggaussian, li2025instag, d3talker}. Methods like TalkingGaussian \cite{li2024talkinggaussian}, InsTaG \cite{li2025instag}, and DEGSTalk \cite{degstalk} further decompose Gaussian fields into separate face and mouth branches to learn movements with different frequencies. Despite improved lip-synchronisation and real-time rendering, these methods are prone to approximation errors introduced by tri-planes, which propagate to the deformation stage. Our work improves on current baselines by modelling speech-driven deformations through learnt per-Gaussian embeddings instead of tri-plane features. 

\subsection{Deformable 3DGS}

General methods for deforming 3D Gaussians are typically aimed at 4D scene reconstruction. Here, the deformation signal is modelled as a temporal embedding conditioned on video frame order. Both Deformable 3D Gaussians \cite{yang2024a} and Dynamic 3D Gaussians \cite{dynamicgaussians} directly deform the Gaussian attributes without using any intermediate representations. 4DGS \cite{4dgs_iclr} combines conditional 3D Gaussians with a marginal 1D Gaussian for time-dependent view synthesis. 4D-GS \cite{Wu_2024_CVPR} decomposes Gaussian centres with a multi-resolution HexPlane \cite{Cao2023HexPlane}, an extension of the tri-plane with an additional time axis, before deformation. FreeTimeGS \cite{wang2025freetimegs} endows each Gaussian with a motion function to predict deformations. E-D3DGS \cite{ed3dgs} defines learnable embeddings for each Gaussian to model time-based deformations. However, temporal embeddings do not encode any acoustic information, and are therefore unsuitable for synthesising motions that arise from speech. In this work, we show how the Gaussian embedding-based deformation setup can be adapted to talking head synthesis.

\section{Method}

\begin{figure*}
  \centering
  \includegraphics[width=0.95\textwidth]{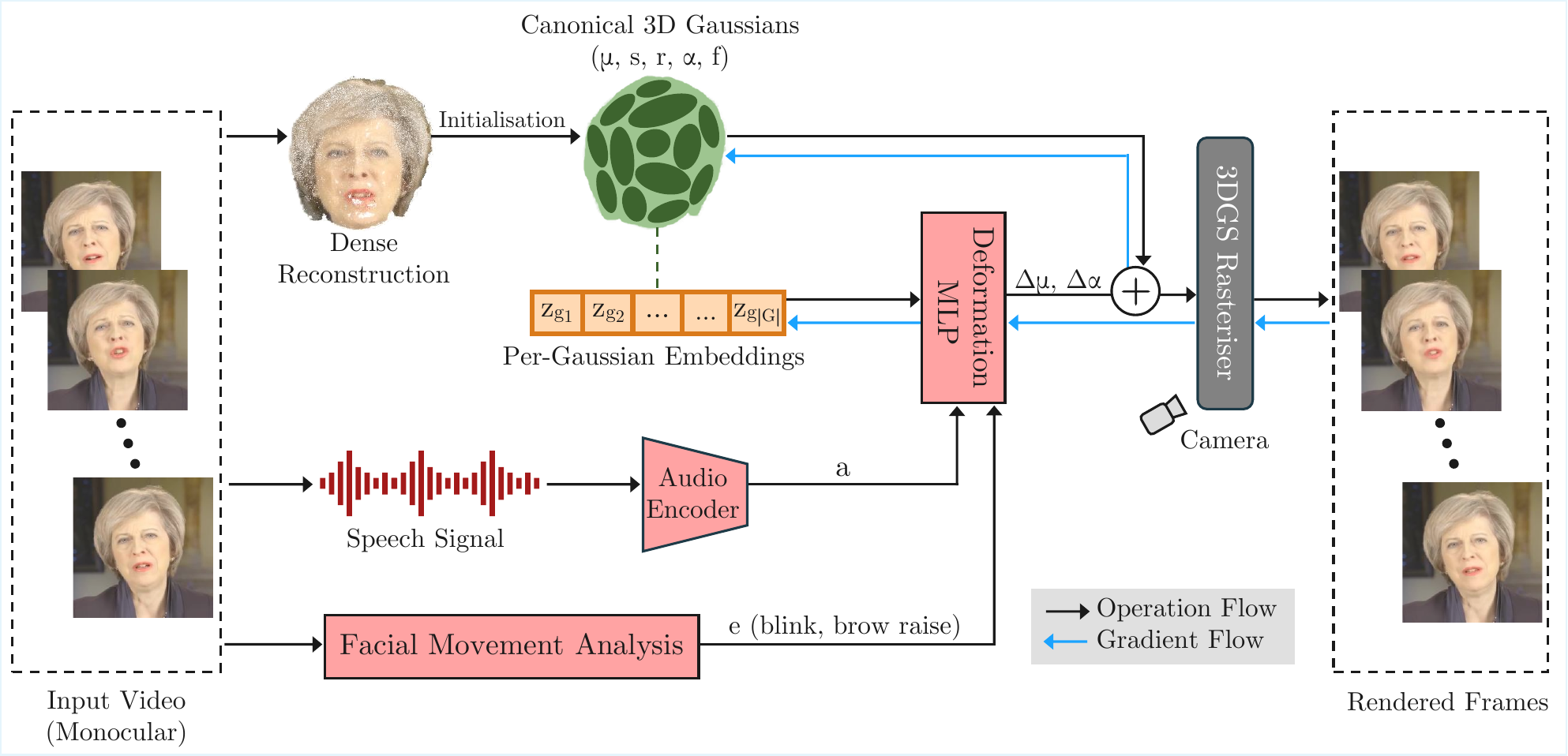}
  \caption{EmbedTalk begins with a talking portrait video. The video frames enable a dense reconstruction of the head that initialises the 3D Gaussians. Each Gaussian is also associated with a learnable embedding $z_g$. For each frame, the corresponding speech signal $a$ and upper-face movements $e$ are fed into the deformation MLP, along with a positional encoding of $z_g$ to predict the Gaussian deformations ($\Delta\mu, \Delta\alpha$). The deformed Gaussians are passed to the rasteriser, along with the viewing direction (camera), to render the head onto onto a combined image containing the torso and scene background.}
  \label{fig:overview}
\end{figure*}

As illustrated in \cref{fig:overview}, EmbedTalk deforms learnable Gaussian embeddings to synthesise an identity-specific talking head from a monocular video and corresponding speech audio. We begin with a brief overview  of 3D Gaussian Splatting (\cref{subsec:3dgs}). Next, we formalise our problem and describe the embedding-based approach to synthesise a talking head from audio (\cref{subsec:3dgs_talking_heads}). Finally, \cref{subsec:training} details the training process. 

\subsection{Preliminary: 3D Gaussian Splatting}
\label{subsec:3dgs}

3DGS \cite{3dgs} optimises a set of anisotropic 3D Gaussians using differentiable tile rasterisation to learn a static 3D scene representation. Each Gaussian is defined using a centre mean $\mu \in \mathbb{R}^3$ and covariance matrix $\Sigma \in \mathbb{R}^{3\times3}$ :
\begin{equation}
    G(x) = e^{\frac{-1}{2}{(x - \mu)}^T\Sigma^{-1}(x - \mu)}
\end{equation}
for a 3D point $x \in \mathbb{R}^3$. The covariance matrix $\Sigma$ is decomposed into a scaling matrix $S$ and a rotation matrix $R$, with scaling factor $s \in \mathbb{R}^3$ and rotation quaternion $q \in \mathbb{R}^4$:
\begin{equation}
    \Sigma = RSS^TR^T
\end{equation}
Each Gaussian also has an opacity value $\alpha \in \mathbb{R}$, and  a colour feature, $f \in  \mathbb{R}^{3(d+1)(d+1)}$, described using a set of spherical harmonics with degree d. To render images, the 3D Gaussians are projected to 2D by calculating the covariance matrix $\Sigma'$ in the camera space using the viewing transformation $W$:
\begin{equation}
    \Sigma' = JW\Sigma W^TJ^T 
\end{equation}
where $J$ is the Jacobian of the affine approximation of the projective transformation. The colour of each pixel is computed by blending all $\mathcal{N}$ depth-ordered Gaussians that overlap the pixel:
\begin{equation}
    C = \sum_{i=1}^{\mathcal{N}} c_i\alpha'_i \prod_{j=1}^{i-1}(1-\alpha'_j)
\end{equation}
where $c_i$ is the colour of each point determined using the colour feature $f_i$ with the viewing direction, and $\alpha'$ is the product of the opacity $\alpha$ with the evaluation of the 2D Gaussian with covariance $\Sigma'$. Cameras calibrated using Structure-from-Motion (SfM) \cite{sfm, colmap} serve as the input, and the 3D Gaussians are initialised using the sparse point cloud created during SfM.\\

\subsection{3DGS for Talking Head Synthesis} 
\label{subsec:3dgs_talking_heads}

Generating talking heads with 3DGS typically requires a talking portrait video $V=\{v_n\}$ consisting of $|V|$ frames, along with corresponding speech signals $A=\{a_n\}$ and optional facial control signals $E=\{e_n\}$. The goal is to construct a canonical 3D representation of the head ($G_{canonical}$), and learn a module to predict deformations to $G_{canonical}$ for each frame based on $a_n$ and $e_n$, where $1 \leq n \leq |V|$. This involves projecting one or more Gaussian primitives to a set of continuous, lower dimensional subspaces. This lower dimensional representation is then combined with the audio embedding, and control signals (if applicable), and reprojected back into the Gaussian space to produce an attribute-wise deformation with respect to the canonical representation. The reprojection is achieved using a decoder multi-layer perceptron (MLP):
\begin{gather}
    G_{canonical} = \{\mu, r, s, f_, \alpha\}\\
    \begin{split}\{\Delta\mu_n, \Delta r_n, \Delta s_n, \Delta f_n, \Delta\alpha_n\} =\\MLP(H(G_{canonical})\oplus a_n \oplus e_n)\end{split}\\
    \begin{split}G^n_{deform} = \{\mu + \Delta\mu_n, r + \Delta r_n, s + \Delta s_n,\\ f + \Delta f_n, \alpha + \Delta\alpha_n\}\end{split}
\end{gather}
where $H$ is a multi-resolution tri-plane encoder \cite{synctalk, cho2024gaussiantalker, li2024talkinggaussian, li2025instag} and $\oplus$ denotes concatenation.

Despite their popularity in radiance field-based talking head generation, tri-plane representations are limited by grid resolution and approximation errors. We therefore do not use tri-planes to drive the Gaussian deformations for our method. Instead, we start with the embedding-based deformation paradigm introduced in E-D3DGS \cite{ed3dgs}, where each Gaussian has a learnable embedding $z_g \in \mathbb{R}^{32}$, in addition to the canonical attributes. E-D3DGS uses temporal embeddings (approximated from frame numbers) to deform Gaussian embeddings for 4D scene reconstruction. To allow deformations to be driven by acoustic features like phonemes, pitch, and amplitude, we replace the temporal embeddings in E-D3DGS with audio embeddings derived from the speech signal. Furthermore, 4D scene reconstruction is dominated by large and (relatively) low-frequency movements, in contrast to portrait lip-sync, where small (relative to the face) and high-frequency mouth movements are the primary source of deformation. To capture these high-frequency details, we apply alternating sine and cosine functions of varying frequencies to the per-Gaussian embeddings on input to the MLP, typically referred to as positional encodings ($\gamma$) \cite{mildenhall2021nerf}. However, our application of positional encodings is not to provide a notion of sequence \cite{transformer_nips}. For EmbedTalk, this mapping into a higher dimensional space enables the embeddings to disentangle motion discontinuous from smooth deformations (for \eg, Gaussians at the lips moving \textit{apart} due to speech, while also shifting \textit{together} laterally due to a head tilt). Additionally, we note the presence of facial movements that are uncorrelated with speech (blinking, brow raise). We provide this information to the deformation MLP through the feature set $E$, comprising six facial action units \cite{ekman1978}. Our embedding-based deformation is given by:

\begin{gather}
    G_{canonical} = \{\mu, r, s, f_, \alpha, z\}\\
    \{\Delta\mu_n, \Delta\alpha_n\} = MLP(\gamma(z)\oplus a_n \oplus e_n)\\
    G^n_{deform} = \{\mu + \Delta\mu_n, r, s, f, \alpha + \Delta\alpha_n\}
\end{gather}

The deformation module for EmbedTalk is a shallow MLP with a separate prediction head for each deformed attribute ($\mu$ and $\alpha$). Unlike contemporary methods that deform all Gaussian attributes \cite{ed3dgs, cho2024gaussiantalker}, or attributes related to size and orientation \cite{degstalk, li2024talkinggaussian, li2025instag}, we choose to deform only position and opacity. This selection is grounded in the fact that facial animation primarily involves changes in motion (head tilting, mouth opening and closing) and visibility (teeth/tongue appearing or disappearing). The facial structure and characteristics (size of nose, distance between eyes) remain unchanged. We validate our selection through experimental variations presented in \cref{subsec:abl_var}.  

\subsection{Training}
\label{subsec:training}

\textbf{Initialisation:} Prior work infers camera poses for 3DGS by fitting a 3D Morphable Model (3DMM) \cite{3dmm} to the input frames \cite{li2024talkinggaussian, cho2024gaussiantalker, degstalk, li2025instag}. The Gaussians are then initialised with a random point cloud \cite{li2024talkinggaussian, degstalk, li2025instag}, or using the coordinates of mesh vertices obtained from the 3DMM fitting \cite{cho2024gaussiantalker}. However, 3DMM fitting is imprecise and often leads to wobbling around the face \cite{gaussianheadtalk}. SfM-based initialisation provides better reconstruction than random initialisation \cite{3dgs}. Consequently, we initialise the Gaussians for EmbedTalk using a dense reconstruction obtained from COLMAP, downsampled to $\leq$ 100K points (to prevent out of memory errors). To quantify the absolute performance gains for our embedding‑based approach, we also report results for 3DMM‑based initialisation. This setup uses a semi-dense point cloud obtained from 3DMM mesh vertices, which is $\sim$2.6x smaller than the COLMAP reconstruction due to the limited number of vertices in the 3DMM model. During training, the Gaussian count is controlled through adaptive densification and pruning strategies \cite{3dgs}. 

\textbf{Rendering:} By default, 3DGS renders Gaussians on a single colour background image, which is unideal for our approach. Following previous work \cite{guo2021adnerf, li2023ernerf}, we render the deformed head onto a combined image containing the torso and scene background using a modified rasteriser \cite{cho2024gaussiantalker}. This prevents artefacts around facial contours.

\textbf{Optimisation:} We jointly optimise the canonical Gaussian attributes, per-Gaussian embeddings, and deformation module to minimise the rendering loss $\mathcal{L}_1$. We also use a perceptual loss, $\mathcal{L}_{LPIPS}$ \cite{lpips}, to capture finer details for the full image as well as the localised mouth region (obtained through facial landmark detection \cite{face_alignment}). Furthermore, we note that nearby Gaussians tend to share similar canonical attributes. It therefore follows that nearby Gaussians should also have similar embeddings to promote motion consistency. Motivated by prior work \cite{dynamicgaussians, ed3dgs}, we apply a local smoothness constraint to encourage similar embeddings for nearby Gaussians, given by:

\begin{equation}
    \mathcal{L}_{emb\_reg} = \frac{1}{k|G|} \sum_{i\in G} \sum_{j\in \text{KNN}_{i;k}} (w_{i,j}\| \mathbf{z}_\text{g}{}_i - \mathbf{z}_\text{g}{}_j \|_2)
\end{equation}

where $k = 20$ is the number of neighbouring Gaussians, $w_{i,j}=e^{(-\lambda_{w} \| \mu_{j} - \mu_{i} \|^2_2)}$ is the weighting factor, and $\lambda_{w}$ is set to 2000 \cite{dynamicgaussians}. To reduce computational overhead, the k nearest neighbours are only computed post densification. Finally, we minimise the mean of the Gaussian opacities to mitigate floaters. The total loss for our setup is given by:

\begin{equation}
\begin{split}
    \mathcal{L}_{total} = \mathcal{L}_{1}(v_n, \hat{v}_n) + \lambda_{face}\mathcal{L}_{LPIPS}(v_n, \hat{v}_n)\\+ \lambda_{mouth}\mathcal{L}_{LPIPS}(v^{m}_n, \hat{v}^{m}_n) + \mathcal{L}_{emb\_reg}\\+ \lambda_{opa}\mathcal{L}_{opacity}
\end{split}
\end{equation}

where $v_n \in V$ represents the ground truth frame, $\hat{v}_n$ is the predicted frame, $v^m_n$ denotes the ground truth frame cropped to the mouth region, and  $\hat{v}^m_n$ is the prediction cropped to the mouth region.

\section{Experiments}
\label{sec:exps}

\textbf{Dataset and Comparisons:} We source five high-definition audio-visual clips from open datasets previously used in similar works \cite{li2024talkinggaussian, hdtf, ye2023geneface}. The clips have an average length of around 6300 frames and are sampled at 25 frames per second (FPS). They comprise three male (Macron, Paul, Obama) and two female (May, Stabenow) identities. Each clip is cropped to centre the portrait, and resized to size 512x512, except the Obama clip, which has size 450x450. We compare our approach with recent 3DGS-based methods: TalkingGaussian \cite{li2024talkinggaussian}, GaussianTalker \cite{cho2024gaussiantalker} and DEGSTalk \cite{degstalk}. To contextualise our work beyond 3DGS-based synthesis, we also compare EmbedTalk with state-of-the-art image-based generative methods: AniTalker \cite{anitalker}, FLOAT \cite{ki2025float}, KDTalker \cite{kdtalker} and Sonic \cite{ji2025sonic}.

\textbf{Implementation:} All 3DGS-based methods, including ours, need identity-specific training. To ensure fair comparison, we use the same train-test split (10:1 ratio) and audio encoder (a pretrained HuBERT model \cite{hsu2021hubert}) for each of these methods. In contrast, generative methods are not tailored to any specific identity. We report their performance using the pretrained weights. Our method is implemented in PyTorch \cite{paszke2019pytorch}. For each identity, we train EmbedTalk for 50,000 iterations using the Adam optimiser \cite{adam}. All other Gaussian-based methods are trained using their recommended configurations. The action units for the expression set $E$ are extracted using OpenFace \cite{Baltrusaitis2016}. For the loss function, we set $\lambda_{face} = 0.01$, $\lambda_{mouth} = 0.002$, and $\lambda_{opa} = 0.0001$. All experiments are conducted on a single NVIDIA L40S 48 GB GPU. The training takes around 1 hour for each identity (\cref{tab:3dgs_compute}). Further specifics are provided in the supplementary material.

\begin{table*}
\resizebox{\textwidth}{!}{
\begin{tabular}{llccccccc}
\specialrule{1.6pt}{1pt}{1pt}
    \textbf{Method} & \textbf{Framework} & \textbf{PSNR$\uparrow$} & \textbf{SSIM$\uparrow$} & \textbf{LPIPS$\downarrow$} & \textbf{LMD$\downarrow$} & \textbf{Sync-C$\uparrow$} & \textbf{FPS (L40S)$\uparrow$} & \textbf{FVMD$\downarrow$}\\ 
    \hline
    Ground Truth & - & $\infty$ & 1 & 0 & 0 & 7.82 $\pm$ 1.53 & - & 0 \\ 
    \hline
    GaussianTalker \cite{cho2024gaussiantalker} & 3DGS & 31.56 $\pm$ 1.95* & 0.92 $\pm$ 0.02* & \cellcolor{third}0.04 $\pm$ 0.01** & 2.76 $\pm$ 0.26 & 6.38 $\pm$ 1.36 & 128 $\pm$ 20*** & 358.02 $\pm$ 117.93* \\ 
    TalkingGaussian \cite{li2024talkinggaussian} & 3DGS & 31.67 $\pm$ 1.59** & 0.92 $\pm$ 0.02* & \cellcolor{second}0.03 $\pm$ 0.01* & \cellcolor{second}2.46 $\pm$ 0.22 & 6.02 $\pm$ 1.33 & \cellcolor{first}303 $\pm$ 26 & \cellcolor{third}347.99 $\pm$ 115.10* \\ 
    DEGSTalk \cite{degstalk} & 3DGS & \cellcolor{third}32.51 $\pm$ 2.61 & \cellcolor{third}0.93 $\pm$ 0.02* & \cellcolor{third}0.04 $\pm$ 0.01 & 2.55 $\pm$ 0.23 & 6.03 $\pm$ 1.40 & \cellcolor{second}296 $\pm$ 8 & 406.06 $\pm$ 89.69** \\ 
    \hline
    KDTalker \cite{kdtalker} & Diffusion & 15.54 $\pm$ 0.99*** & 0.68 $\pm$ 0.05*** & 0.39 $\pm$ 0.05*** & 7.05 $\pm$ 1.15*** & \cellcolor{third}6.64 $\pm$ 1.68 & 28 $\pm$ 1*** & 2974.68 $\pm$ 579.42*** \\ 
    Sonic \cite{ji2025sonic} & Diffusion & 18.90 $\pm$ 1.56*** & 0.74 $\pm$ 0.04*** & 0.22 $\pm$ 0.07** & 5.04 $\pm$ 0.99** & \cellcolor{first}8.09 $\pm$ 1.16 & $<$ 2*** & 2313.82 $\pm$ 343.82*** \\ 
    AniTalker \cite{anitalker} & Diffusion & 17.12 $\pm$ 1.08*** & 0.72 $\pm$ 0.03*** & 0.30 $\pm$ 0.04*** & 5.43 $\pm$ 0.72*** & 6.02 $\pm$ 1.65 & 30 $\pm$ 2*** & 2445.56 $\pm$ 391.26*** \\ 
    \hline
    FLOAT \cite{ki2025float} & Flow Matching & 18.00 $\pm$ 1.30*** & 0.72 $\pm$ 0.03*** & 0.26 $\pm$ 0.05*** & 5.19 $\pm$ 0.92** & \cellcolor{second}7.10 $\pm$ 1.15 & 32 $\pm$ 2*** & 2515.67 $\pm$ 486.89*** \\ 
    \hline
    \textbf{EmbedTalk (3DMM)} & 3DGS & \cellcolor{second}33.20 $\pm$ 1.53 & \cellcolor{second}0.94 $\pm$ 0.02 & \cellcolor{second}0.03 $\pm$ 0.01* & \cellcolor{third}2.47 $\pm$ 0.21 & 6.30 $\pm$ 1.42 & 270 $\pm$ 7 & \cellcolor{second}292.27 $\pm$ 68.42* \\
    \textbf{EmbedTalk (COLMAP)} & 3DGS & \cellcolor{first}35.19 $\pm$ 1.39  & \cellcolor{first}0.96 $\pm$ 0.01 & \cellcolor{first}0.02 $\pm$ 0.00 & \cellcolor{first}2.44 $\pm$ 0.18 & 6.52 $\pm$ 1.51 & \cellcolor{third}294 $\pm$ 5 & \cellcolor{first}147.38 $\pm$ 36.12 \\ 
    \specialrule{1.6pt}{1pt}{1pt}
\end{tabular}}
\caption{Results (mean $\pm$ sd) for the \textbf{self-driven} setting, with the top three averages in red (first), orange (second) and yellow (third). EmbedTalk achieves the best rendering quality, (personalised) lip-sync, and motion consistency, along with a high inference speed. Statistical significance relative to EmbedTalk (COLMAP) is assessed using Welch’s t-test, with *, **, and *** denoting $p<$0.05, $p<$0.01, and $p<0.001$ respectively. Note that there are no statistically significant differences for the Sync-C metric.}
\label{tab:self_driven}
\end{table*}

\subsection{Quantitative Evaluation}

\begin{table}
\resizebox{\columnwidth}{!}{
\setlength{\tabcolsep}{2.5mm}
\begin{tabular}{llccc}
\hline
\multirow{2}{*}{\textbf{Method}} & \multirow{2}{*}{\textbf{Framework}}  & \multicolumn{3}{c}{\textbf{Sync-C$\uparrow$}} \\ \cline{3-5} 
 &  & \textbf{Cross-Gender} & \textbf{Cross-Lingual} & \textbf{Overall} \\ \hline
GaussianTalker \cite{cho2024gaussiantalker} & 3DGS & 6.23 $\pm$ 0.74 & 2.38 $\pm$ 0.44** & 5.51 $\pm$ 1.84 \\ 
TalkingGaussian \cite{li2024talkinggaussian} & 3DGS & 5.26 $\pm$ 1.09 & 3.56 $\pm$ 0.40 & 4.83 $\pm$ 1.28 \\ 
DEGSTalk \cite{degstalk} & 3DGS & 2.34 $\pm$ 1.74** & 1.47 $\pm$ 0.33*** & 2.29 $\pm$ 1.88* \\ \hline
KDTalker \cite{kdtalker} & Diffusion & \cellcolor{second}7.21 $\pm$ 1.17 & \cellcolor{third}7.35 $\pm$ 0.49** & \cellcolor{second}7.07 $\pm$ 0.43\\ 
Sonic \cite{ji2025sonic} & Diffusion & \cellcolor{first}8.14 $\pm$ 0.65* & \cellcolor{first}9.34 $\pm$ 0.74** & \cellcolor{first}8.30 $\pm$ 0.66*\\ 
AniTalker \cite{anitalker} & Diffusion & \cellcolor{third}6.72 $\pm$ 1.62 & \cellcolor{second}7.48 $\pm$ 0.72* & 6.60 $\pm$ 0.96 \\ \hline
FLOAT \cite{ki2025float} & Flow Matching & 6.44 $\pm$ 0.72 & 6.88 $\pm$ 1.21 & \cellcolor{third}6.64 $\pm$ 0.29 \\ \hline
\textbf{EmbedTalk (COLMAP)} & 3DGS & 6.02 $\pm$ 1.67 & 4.28 $\pm$ 0.17 & 6.01 $\pm$ 1.55 \\ \hline
\end{tabular}}
\caption{Results (mean $\pm$ sd) for the \textbf{cross-driven} setting, with the top three averages in red (first), orange (second) and yellow (third). Generative methods dominate due to exaggerated mouth movements. Statistical significance relative to EmbedTalk is assessed using Welch’s t-test, with *, **, and *** denoting $p<$0.05, $p<$0.01, and $p<0.001$ respectively.}
\label{tab:cross_driven}
\end{table}

We use PSNR (Peak Signal-to-Noise Ratio), SSIM (Structural Similarity Index Measure) and LPIPS (Learned Perceptual Image Patch Similarity) to measure rendering quality. For audio-visual alignment, we employ Landmark Distance (LMD) \cite{Chen2018LipMG}, which measures the distance between real and predicted mouth landmarks, and the confidence scores (Sync-C) from the SyncNet model \cite{Chung16a}. Furthermore, we leverage Fréchet Video Motion Distance (FVMD) \cite{fvmd} to evaluate the motion consistency of the generated videos. FVMD computes the Fréchet distance between motion features of generated and ground-truth videos, derived using target point velocities and accelerations. We also report inference speed by measuring FPS on an enterprise GPU that can accommodate all models. All metrics are computed over the full frame, including the torso and background.

Videos are rendered under two different settings. The first is the \textbf{self-driven} setting, where videos generated for an identity are driven by unseen speech sourced from the same identity (the 10:1 train-test split). The second is the  \textbf{cross-driven} setting, where videos are generated using: (1) speech from other identities, and (2) synthetic speech created using a text-to-speech (TTS) model. For this, we source human speech from the HDTF (High-Definition Talking Face) dataset \cite{hdtf}, and generate AI samples using the voices \textit{Adaline, Autumn, Hale, Iron Rose, Jamie} and \textit{Michael} from ElevenLabs \cite{tts}. The text for the TTS model is taken from the works of William Shakespeare.

\begin{figure}
\centering
\subfloat[EmbedTalk (3DMM)]{%
\resizebox*{0.4\columnwidth}{0.4\columnwidth}{\includegraphics{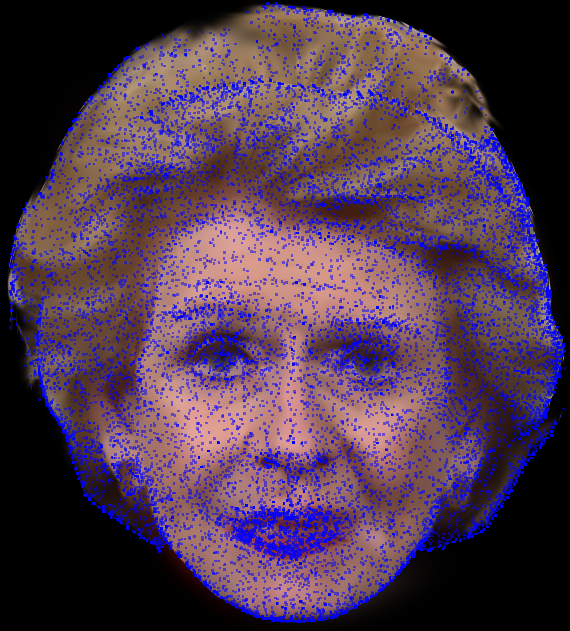}}}\hspace{1cm}
\subfloat[DEGSTalk (3DMM)]{%
\resizebox*{0.4\columnwidth}{0.4\columnwidth}{\includegraphics{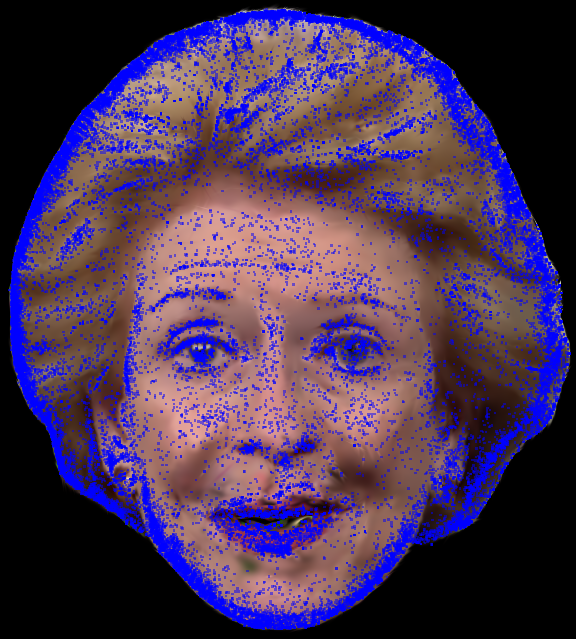}}}
\caption{Gaussian centres (blue dots) for the head reconstructions obtained with (a) EmbedTalk and (b) DEGSTalk, visualised using the SuperSplat Editor. Note the high density of Gaussians around the boundary of the head in DEGSTalk.} 
\label{fig:3dmm_init_points}
\end{figure}

\cref{tab:self_driven} reports performance for all methods under the \textbf{self-driven} setting. EmbedTalk (COLMAP) performs best on all rendering metrics and achieves the most temporally consistent motion. Our method also performs best for identity-specific lip synchronisation (LMD) and has the highest Sync-C score among  3DGS-based methods, enabled by positional encodings that capture high-frequency details. The performance gains for EmbedTalk (3DMM) are smaller due to the unstable sparse initialisation. But while wobbling around the head boundary persists, our method reduces overall shakiness, as evidenced by the density of white pixels in the upper head region of the renderings shown in \cref{fig:wobble}. Positional encodings combined with our prune-first optimisation strategy encourage sparsity in regions with low frequency movements, such as the boundary of the head. Consequently, our model has fewer Gaussians in these regions that can learn any noise introduced by inaccurate facial tracking, as shown in \cref{fig:3dmm_init_points}. These findings suggest that our improvements stem from both stronger initialisation and the proposed architectural choices, rather than from the head pose estimation method alone. 

Generative methods fare poorly on rendering metrics due to lack of pose information, and are limited by inference speed. However, they achieve high Sync-C scores due to the production of exaggerated mouth movements, with Sonic's score being even higher than the ground truth. For the \textbf{cross-driven} setting, we only report Sync-C due to lack of ground truth data (\cref{tab:cross_driven}). We further partition these results by gender (model trained on male/female identity, driven by female/male speech) and language (model trained on one language, driven by speech from another language). While partitioned results are presented for all methods, it is worth noting that cross-gender and cross-lingual settings only apply to identity-specific training regimes (3DGS). Here, generative methods again dominate the Sync-C scores, but our method has the highest overall Sync-C among the Gaussian methods. 
We also observe greater variance across identities for 3DGS-based methods than for generative methods. This can be attributed to the personalised talking heads generated by radiance fields, while generative models tend to produce generic facial motions across all identities.
Interestingly, cross-lingual and cross-gender scores tend to be higher than the overall scores in many instances. 

For the Gaussian methods, we additionally report training time, model size and FPS with a laptop GPU in \cref{tab:3dgs_compute}. Since EmbedTalk does not require a tri-plane encoder, our models are around 2x to 6x smaller than methods which use tri-planes. The inference speed-up provided through this elimination is best realised on the laptop setup, with EmbedTalk being nearly twice as fast as other methods. However, our training takes longer since we begin with a dense point cloud and focus on pruning Gaussians during training, as opposed to growing Gaussians from sparse or randomly initialised point clouds. The final Gaussian counts for each identity are given in the supplementary material. 

\begin{table}
\resizebox{\columnwidth}{!}{
\begin{tabular}{llccc}
\specialrule{1.6pt}{1pt}{1pt}
    \textbf{Method} & \textbf{Deformation Field} & \textbf{Training Time $\downarrow$} & \textbf{FPS (RTX 2060) $\uparrow$} & \textbf{Model Size $\downarrow$}\\ 
    \hline
    GaussianTalker \cite{cho2024gaussiantalker} & Tri-plane & 03:06:19 & 33 & \cellcolor{second}19.51 MB  \\ \hline
    TalkingGaussian \cite{li2024talkinggaussian} & Tri-plane & \cellcolor{first}00:27:50 & \cellcolor{second}38 & \cellcolor{third}27.08 MB  \\ \hline
    DEGSTalk \cite{degstalk} & Tri-plane + Embeddings & \cellcolor{second}00:37:26 & \cellcolor{third}37 &58.69 MB \\ \hline
    \textbf{EmbedTalk (COLMAP)} & Embeddings & \cellcolor{third}1:01:49 & \cellcolor{first}61  & \cellcolor{first}10.20 MB \\ 
    \specialrule{1.6pt}{1pt}{1pt}
\end{tabular}}
\caption{Computational costs for the 3DGS-based methods. EmbedTalk yields compact models that achieve over 60 FPS on a laptop (RTX 2060 6GB) GPU.}
\label{tab:3dgs_compute}
\end{table}

\subsection{Qualitative Evaluation}

\begin{figure*}
  \centering
  \includegraphics[width=0.75\textwidth]{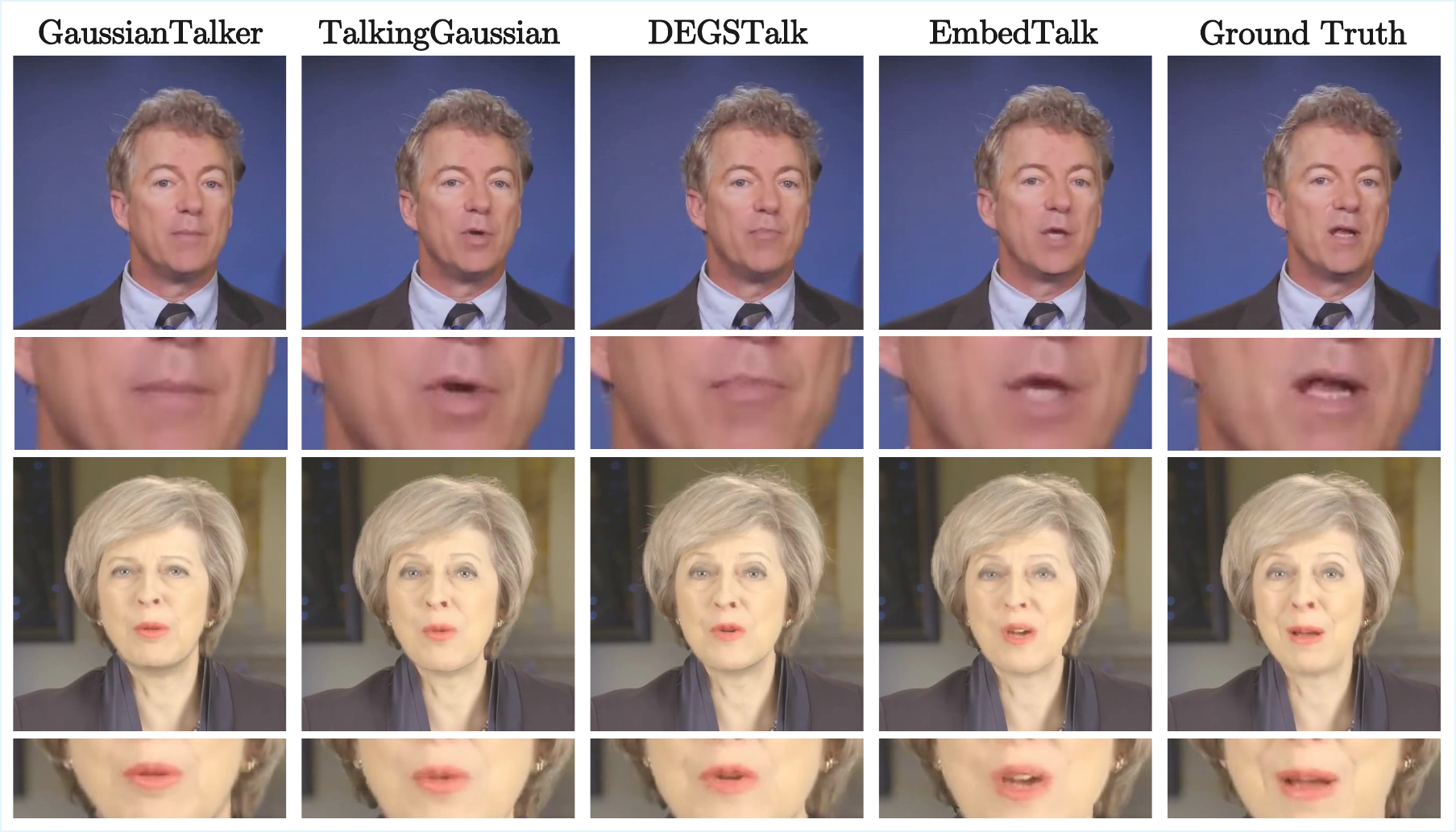}
  \caption{Qualitative comparison with recent 3DGS-based works. EmbedTalk reconstructs mouth openings more faithfully than other methods.}
  \label{fig:comp_3dgs}
\end{figure*}

\begin{figure}
  \centering
  \includegraphics[width=0.95\columnwidth]{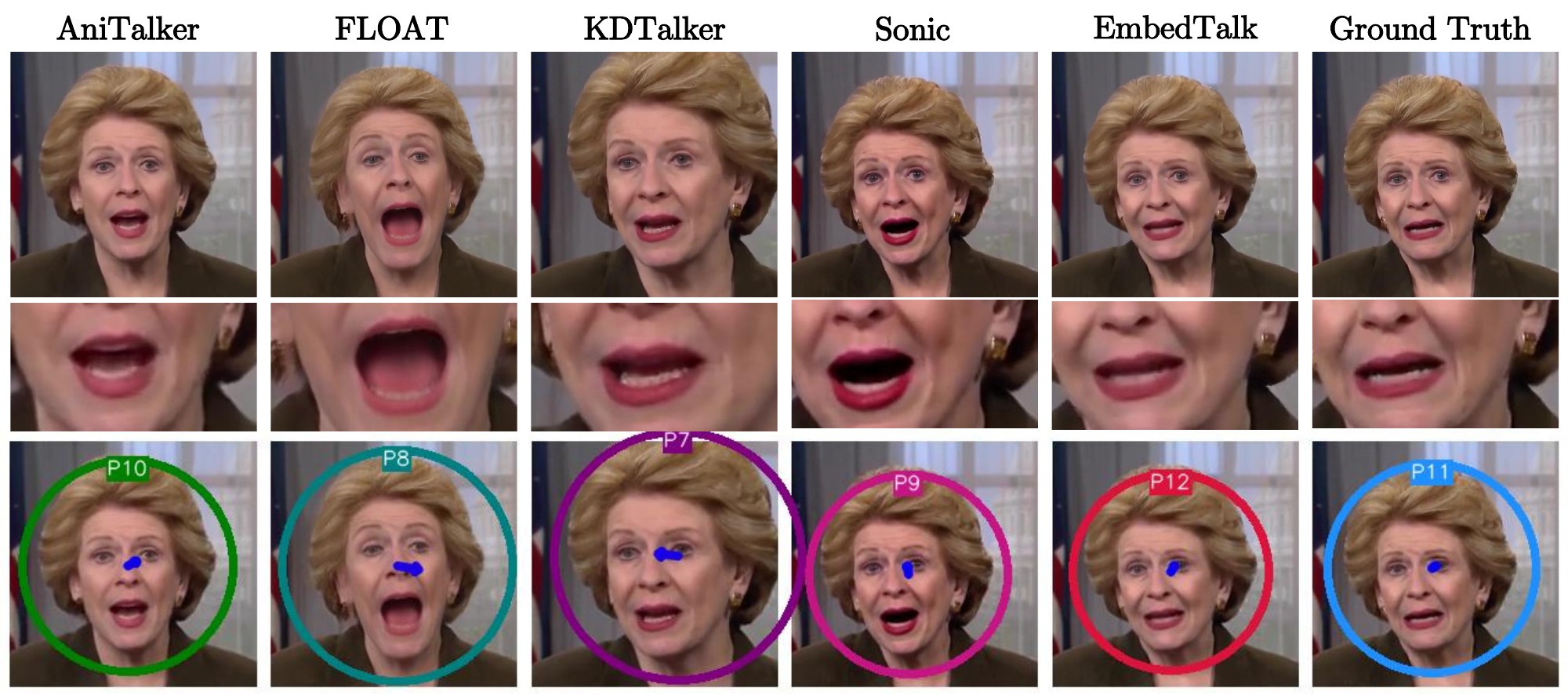}
  \caption{Qualitative comparison with generative methods. Despite accurate lip sync, generative models produce exaggerated movements that reduce realism. Some methods also redirect the speaker’s gaze (blue arrow) away from the camera.}
  \label{fig:comp_generative}
\end{figure}

\begin{figure}
  \centering
  \includegraphics[width=0.95\columnwidth]{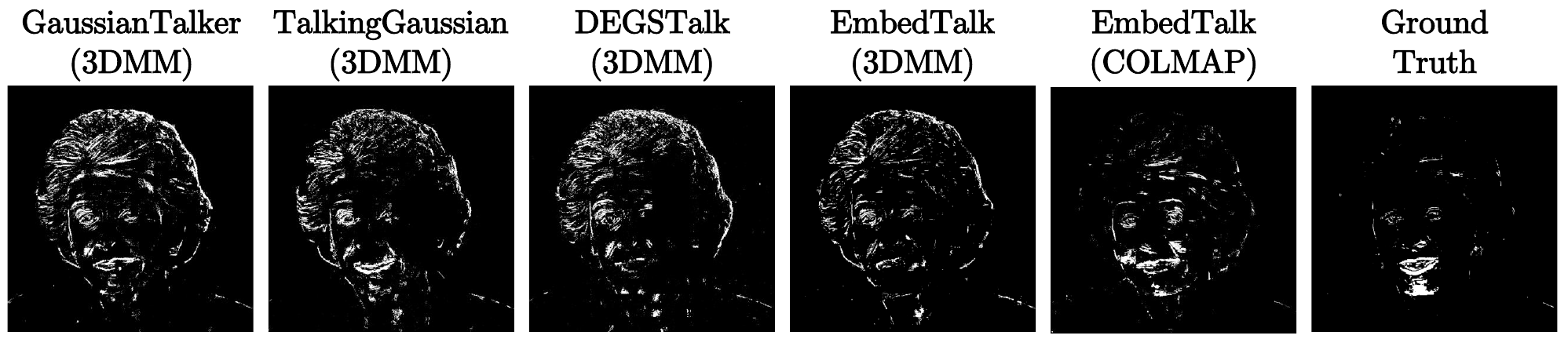}
  \caption{Consecutive differences accumulated over 20 frames. White pixels in the upper head region indicate temporal flickering, with higher intensities corresponding to larger instabilities.}
  \label{fig:wobble}
\end{figure}

To qualitatively assess the rendering quality and audio-visual alignment, we present rendered and ground truth frames from the self-driven setting. A video file showing comparisons among renderings is also available on the project page. \cref{fig:embedtalk_teaser} and \cref{fig:comp_3dgs} provide comparisons with 3DGS-based works. \cref{fig:comp_generative} shows comparisons with generative methods. Among Gaussian Splatting-based approaches, EmbedTalk generates narrow mouth openings faithfully, while other methods often default to a closed-mouth state. For generative models, the mouth movements are accurate but tend to be uncharacteristically large, especially for FLOAT and Sonic, leading to reduced realism. KDTalker and FLOAT also tend to produce frames where the speaker's gaze is directed away from the camera, which may reduce applicability in interactive settings. To highlight this issue, we use the Gaze Transformer model \cite{gaze3d} to estimate the gaze direction for generated frames, denoted by the blue arrows in the bottom row of \cref{fig:comp_generative}. We also illustrate the wobbling effect, mentioned earlier, in 3DGS-based works through accumulated differences between consecutive frames (over a 20 frame interval). The white lines in the upper region of the heads in \cref{fig:wobble} indicate temporal flickering in the renderings for GaussianTalker, TalkingGaussian and DEGSTalk. This issue persists in EmbedTalk (3DMM), but with less intensity, as evidenced by the lower concentration of white pixels in the upper head region. EmbedTalk (COLMAP) produces the most stable rendering that aligns closely with the ground truth.

\begin{table}
\resizebox{\columnwidth}{!}{
\begin{tabular}{lllccccl}
\hline
\multicolumn{2}{l}{\textbf{Experiment Configuration}} & \textbf{PSNR$\uparrow$} & \textbf{SSIM$\uparrow$} & \textbf{LPIPS$\downarrow$} & \textbf{LMD$\downarrow$} & \textbf{Sync-C$\uparrow$} & \textbf{FVMD$\downarrow$} \\ \hline
\multirow{2}{*}{Ablations} & w/o pos. enc. & 35.22 & 0.96 & 0.02 & 2.58 & 5.57 & 157.00 \\ \cline{2-8} 
 & w/o emb. reg. & 35.11 & 0.96 & 0.02 & \cellcolor{second}2.51 & \cellcolor{third}5.90 & 229.73 \\ \hline
\multirow{2}{*}{$z_g$ dimension} & 16 & 35.04 & 0.96 & 0.02 & 2.63 & 5.09 & 231.20 \\ \cline{2-8} 
 & 64 & 34.66 & 0.96 & 0.02 & 2.73 & 5.76 & 158.47 \\ \hline
\multirow{4}{*}{\begin{tabular}[c]{@{}l@{}}Deformed \\ Attributes\\ $(\Delta\mu,$)\end{tabular}} & $\Delta\alpha, \Delta f$ & 34.96 & 0.96 & 0.02 & \cellcolor{third}2.53 & \cellcolor{second}6.38 & \cellcolor{second}146.85 \\ \cline{2-8} 
 & $\Delta r, \Delta s$  &  34.51 & 0.96 & 0.02 & 3.10 & 2.98 & 260.67 \\ \cline{2-8} 
 & $\Delta r, \Delta s, \Delta f$  & 34.66 & 0.96 & 0.02 & 2.86 & 3.03 & 251.76\\ \cline{2-8} 
 & $\Delta r, \Delta s, \Delta\alpha, \Delta f$ & 34.94 & 0.96 & 0.02 & 2.70 & 5.45 & \cellcolor{first}146.64 \\ \hline
\multicolumn{2}{l}{\textbf{Ours ($z_g=32, \Delta\mu, \Delta\alpha$)}} & 35.19 & 0.96 & 0.02  & \cellcolor{first}2.44 & \cellcolor{first}6.52 & \cellcolor{third}147.38 \\ \hline
\end{tabular}}
\caption{Results for Ablations and Experimental Variations to validate the design choices for EmbedTalk. Due to negligible differences and multiple ties, we do not indicate the top three values in the PSNR, SSIM, and LPIPS columns.}
\label{tab:ablation}
\end{table}

\begin{figure}[h]
\centering
\subfloat[Video Realness]{%
\resizebox*{\columnwidth}{!}{\includegraphics{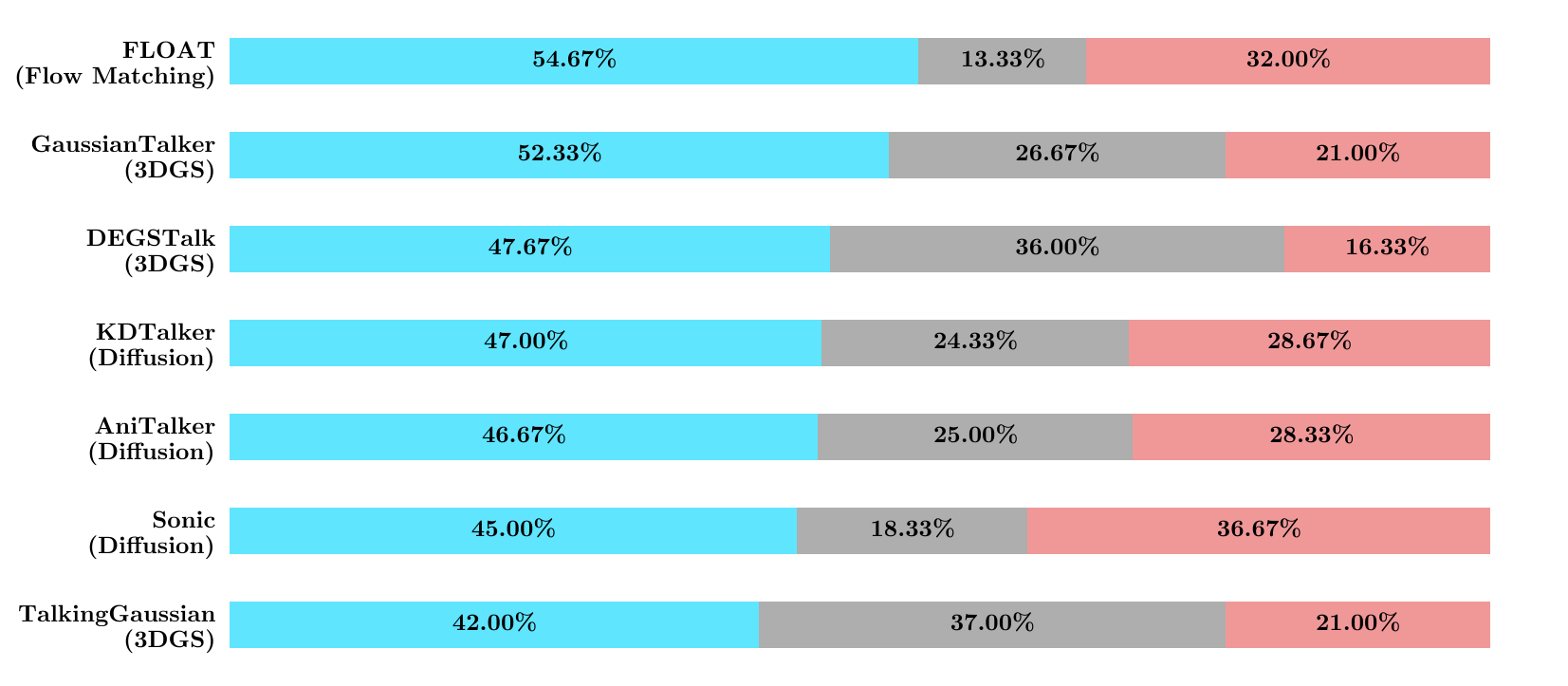}}}\hfill
\subfloat[Lip Synchronisation]{%
\resizebox*{\columnwidth}{!}{\includegraphics{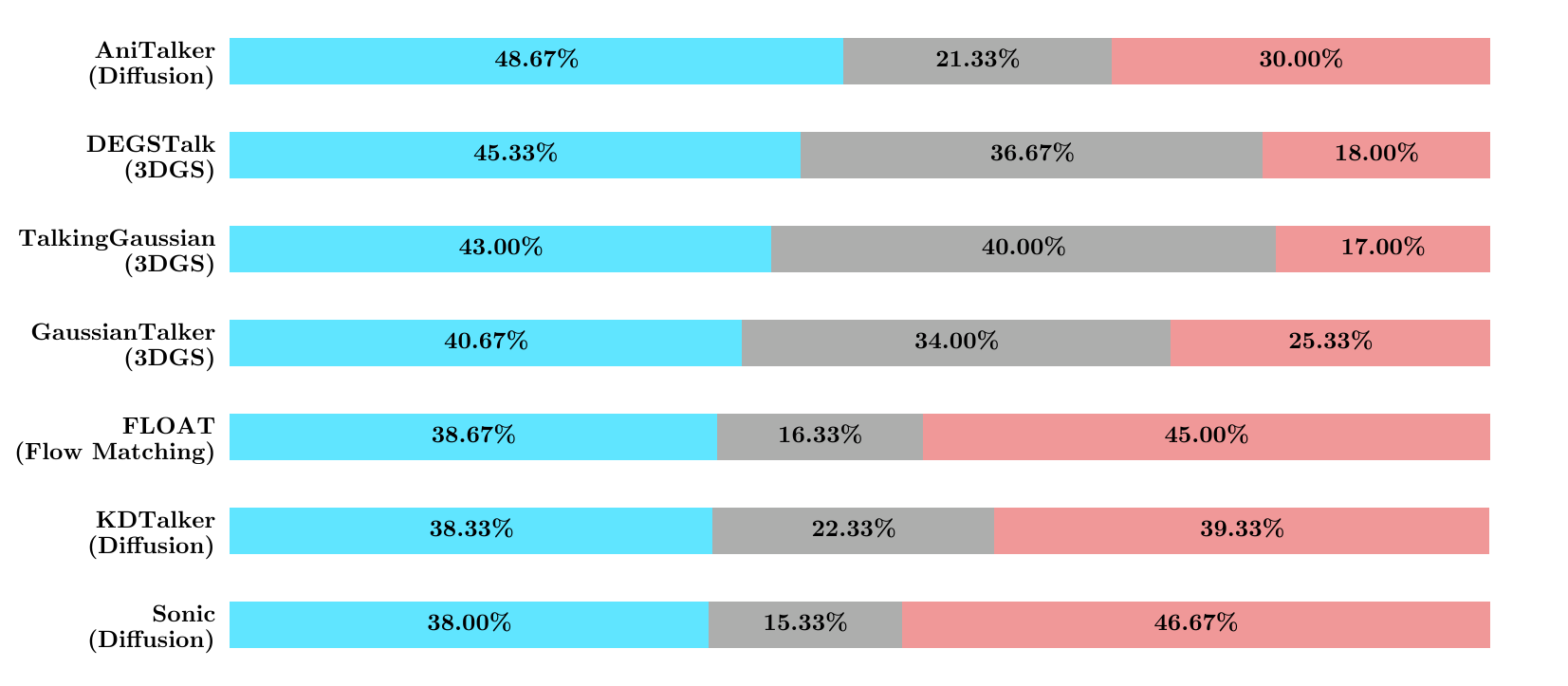}}}\hfill
\subfloat[Image Quality]{%
\resizebox*{\columnwidth}{!}{\includegraphics{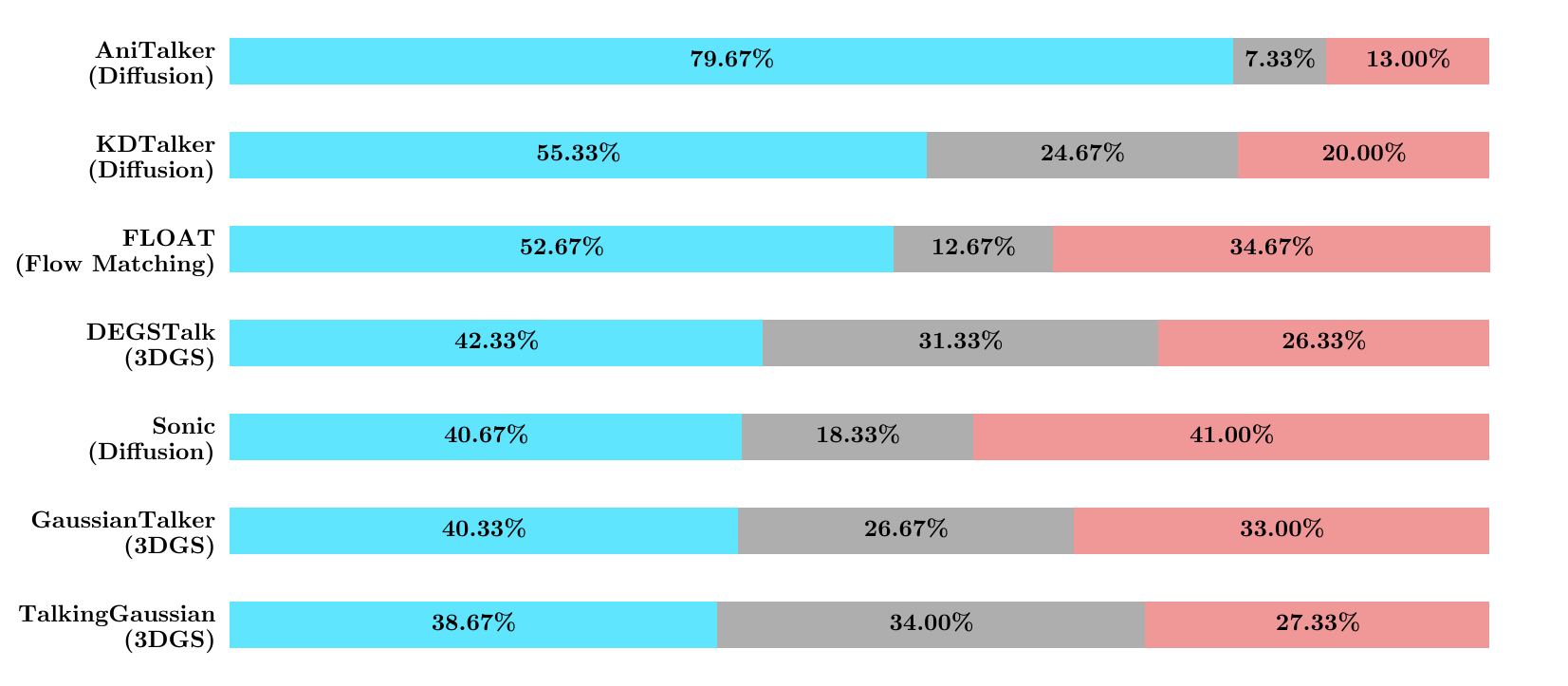}}}\hfill
\subfloat{%
\resizebox*{\columnwidth}{!}{\includegraphics{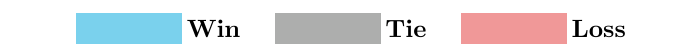}}}
\caption{Results of the user study showing win rates for EmbedTalk in paired setups. Assessment criteria include (a) Video Realness, (b) Lip Synchronisation, and (c) Image Quality.} 
\label{fig:human_eval}
\end{figure}

Beyond visual comparisons, we also conducted a user study to assess the generated videos on three aspects: Image Quality, Lip Synchronisation, and Video Realness. Three videos for each of the five identities (one self-driven and two cross-driven) were paired with corresponding videos generated using the seven comparative baselines. We recruited 20 assessors through the Prolific platform \cite{prolific}. The participants were shown the paired videos and asked to select which video was better in terms of the evaluation aspect, the choices being `left', `right' or `both are the same'. Participants were blinded to the methods used to generate the videos for any given pair. They were asked to focus on sharpness of video frames and preservation of fine details for image quality, and audio-visual alignment for lip-sync. For video realness, the objective was to select the video that seemed less like it had been AI-generated. The protocol for evaluation was reviewed and approved by the Faculty Research Ethics Committee for Engineering and Physical Sciences at the University of Leeds (EPS FREC - 2025 3516-5033). Participants were compensated using the national living wage as the rate, scaled pro rata for the time involved. \cref{fig:human_eval} shows the results of the user study for each aspect, ordered by the win rate (where a win is a preference for our method). EmbedTalk scores high on realism and image quality, whilst being marginally worse than generative methods for lip-sync, owing to their highly pronounced lip movements. We also note frequent ties in paired comparisons with other 3DGS-based methods, which is expected, given the similarity of approach.

\subsection{Ablations and Experiment Variations}
\label{subsec:abl_var}

To validate the efficacy of our design choices, we conduct ablations and experimental variations under the self-driven setting. For ablations, we report performance in the absence of (a) positional encodings and (b) the local smoothness constraint ($\mathcal{L}_{emb\_reg}$). For experimental variations, we present metrics for different embedding sizes ($dim(z_g) = {16, 64}$) and by varying the Gaussian attributes deformed. \cref{tab:ablation} presents these additional experiments. There are negligible differences and multiple ties in the rendering metrics, since the mouth occupies a small portion of the full frame. Consequently, we have not indicated the top three in the PSNR, SSIM and LPIPS columns. 

Our design choices are most beneficial in promoting audio-visual alignment and motion consistency, as evidenced by the LMD, Sync-C, and FVMD values. The local smoothness constraint and our choice to focus on deforming Gaussian position and visibility are especially useful for encouraging smooth motion and high lip synchronisation. We find the deformation of opacity and the colour feature ($\Delta \alpha, \Delta f$ setting) to be the most promising alternative to our setup, albeit with additional predictive overhead.

\section{Conclusion}

We present EmbedTalk, a new method that shows how learnt embeddings can be leveraged to deform 3D Gaussians for talking head synthesis. Results across quantitative and qualitative setups demonstrate that our approach produces high quality talking heads with accurate and realistic lip synchronisation. Furthermore, we show that replacing tri-planes significantly reduces memory usage while improving inference speed on a laptop GPU.

We note some limitations of our work. Although EmbedTalk improves on current baselines, its reconstructions are not perfect (bottom row of \cref{fig:comp_3dgs}). Future work may explore alternative representations beyond tri-plane features and embeddings. Our generated videos are also limited to neutral vocal tone and expression due to the nature of our training data. While we anticipate that EmbedTalk would generalise to new videos with diverse emotions and expressions, this remains to be experimentally validated. Finally, we note that other 3DGS-based works are also likely to improve from a dense point cloud initialisation. However, the optimisation for these setups is non-trivial. We chose to leave these investigations to future efforts.

We hope that our work will support research and development in creative industries. Nonetheless, our method carries potential for misuse through generation of deepfakes for identity theft or misinformation. We advocate for the use of explicit labelling and watermarking techniques to help detect synthetic content. 

\section*{Acknowledgements}
AS is supported by a PhD studentship that is funded by UK Research and Innovation (CDT Grant Reference: EP/S024336/1). This work was undertaken on the Aire HPC system at the University of Leeds, UK. We are grateful to Professor Vania Dimitrova for helpful comments on the presentation of results.

{
    \small
    \bibliographystyle{ieeenat_fullname}
    \bibliography{main}
}

\end{document}